\documentclass[10pt,twocolumn,letterpaper]{article}

\usepackage[font=small]{caption}
\usepackage{cvpr}
\usepackage{times}
\usepackage{epsfig}
\usepackage{graphicx}
\usepackage{amsmath}
\usepackage{amssymb}
\usepackage{svg}

\let\vec\mathbf

\usepackage[breaklinks=true,bookmarks=false]{hyperref}

\cvprfinalcopy 

\def\cvprPaperID{****} 

\begin{document}

\title{Understanding Art through Multi-Modal Retrieval in Paintings}

\author{Noa Garcia \quad Benjamin Renoust \quad Yuta Nakashima\\
Institute for Datability Science\\
Osaka University, Japan\\
{\tt\small \{noagarcia, renoust, n-yuta\}@ids.osaka-u.ac.jp}
}

\maketitle

\begin{abstract}
In computer vision, visual arts are often studied from a purely aesthetics perspective, mostly by analysing the visual appearance of an artistic reproduction to infer its style, its author, or its representative features. In this work, however, we explore art from both a visual and a language perspective. Our aim is to bridge the gap between the visual appearance of an artwork and its underlying meaning, by jointly analysing its aesthetics and its semantics. We introduce the use of multi-modal techniques in the field of automatic art analysis by 1) collecting a multi-modal dataset with fine-art paintings and comments, and 2) exploring robust visual and textual representations in artistic images.
\end{abstract}

\section{Introduction}

The large-scale digitisation of artworks from collections all over the world has opened the opportunity to study art from a computer vision perspective, by building tools to help in the conservation and dissemination of cultural heritage. Some of the most promising work on this direction involves the automatic analysis of paintings, in which computer vision techniques are applied to study the content \cite{crowley2016art}, the style \cite{collomosse2017sketching,sanakoyeu2018style}, or to classify the attributes \cite{mensink2014rijksmuseum,mao2017deepart} of a specific piece of art. In this way, art has been mostly studied from a visual perspective \cite{Bar2014ClassificationOA,karayev2014recognizing,saleh2016large,Tan2016CeciNP,ma2017part}, and less attention has been paid to automatically analyse the underlying meaning of each painting. In this work, we aim to bridge the gap between the visual analysis and the high-level understanding of art, by proposing robust language and vision representations for multi-modal retrieval in paintings.

\begin{figure}
\centering
\includegraphics[width = 0.48\textwidth]{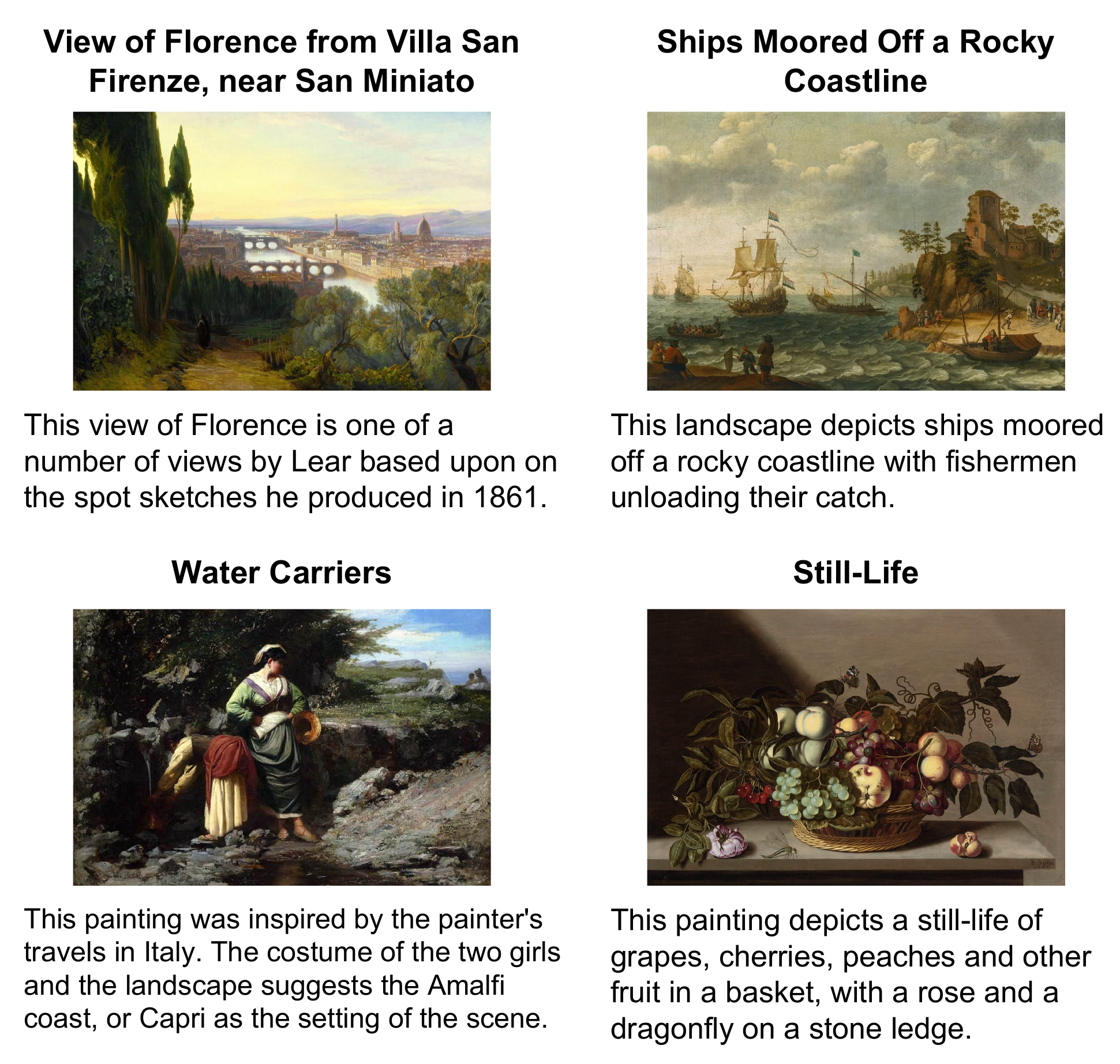}
\caption{Examples of paintings and comments in SemArt dataset.} 
\label{fig:examples}
\end{figure}

We first introduce a multi-modal dataset for visual arts, in which each image of a painting is associated with an artistic comment (Figure \ref{fig:examples}). Differently from multi-modal datasets in natural images, such as VQA \cite{VQA}, Visual Genome \cite{krishna2017visual}, and MS-COCO \cite{Lin2014}, the interpretation of art is strongly related to the artistic context of each artwork. This peculiarity is observed in the proposed dataset both in terms of images, through the use of style and composition, and language, through the use of references. 

To leverage these differences and study art from a semantics perspective, we propose to enhance robust visual and language representations with artistic attributes. The enhanced representations are projected into a multi-modal artistic space in which image and text coexist. By fine-tuning the multi-modal representations in the art domain, paintings and comments that are semantically similar are represented closer than dissimilar samples.

The quality of the proposed multi-modal artistic space is evaluated as a retrieval task, in which given a painting image, the most representative comment from the collection must be found, and vice-versa. Multi-modal retrieval allows us to discriminate whether the language and visual representations capture the sufficient artistic insights to match corresponding paintings and comments together. In the evaluation, our method achieves results only 0.059 below human accuracy.



\begin{figure*}
\centering
\includegraphics[width = \textwidth]{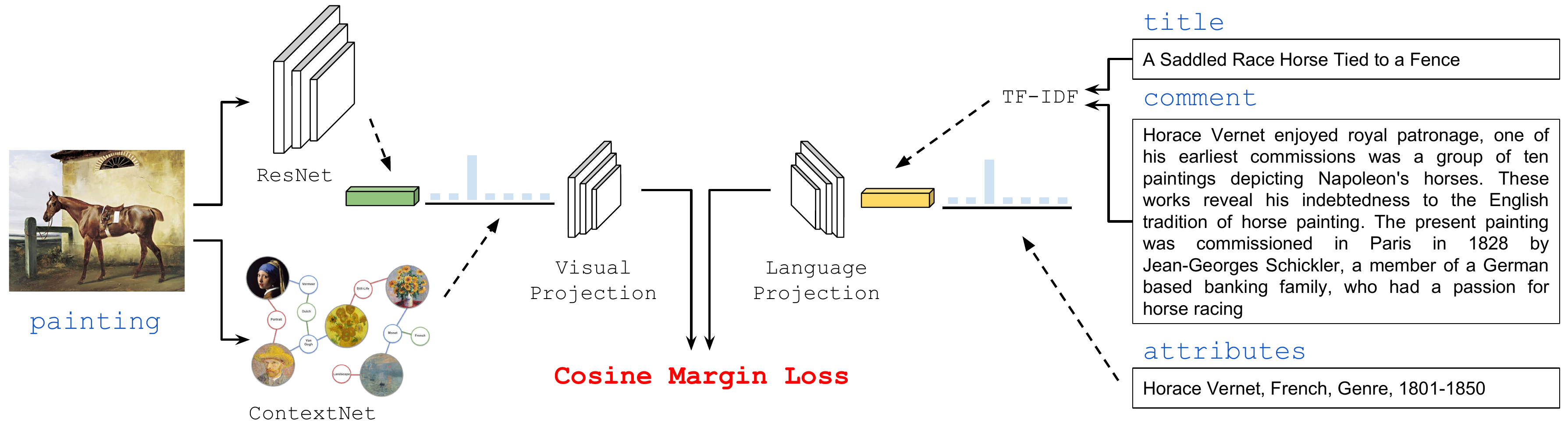}
\caption{Proposed visual and language representations for multi-modal retrieval in art.} 
\label{fig:model}
\end{figure*}

\section{SemArt Dataset}
\label{sec:dataset}
Existing datasets in art analysis, such as PRINTART \cite{carneiro2012artistic}, Painting-91 \cite{khan2014painting}, Rijksmuseum \cite{mensink2014rijksmuseum} or Art500k \cite{mao2017deepart}, are mostly annotated with attribute labels, such as author, style, or timeframe. Although this information is crucial in the analysis of visual arts, it does not provide enough insights for understanding the high-level semantics of fine-art paintings. To jointly study language and vision in art, we introduce SemArt\footnote{Available at \href{http://noagarciad.com/SemArt/}{http://noagarciad.com/SemArt/}}, a dataset for semantic art understanding. 

SemArt contains 21,384 reproductions of European paintings collected from the Web Gallery of Art\footnote{\href{https://www.wga.hu/}{https://www.wga.hu/}}, randomly split into training, validation, and test sets with 19,244, 1,069 and 1,069 samples, respectively. Each image is annotated with its main attributes -- author, title, date, technique, type, school and timeframe\footnote{Periods of 50 years evenly distributed between 801 and 1900.} -- and with a natural language comment. Interestingly, comments involve not only a description of the elements in the scene but also references to its technique, author or context. Some examples are shown in Figure \ref{fig:examples}, and a complete analysis of the dataset can be found in \cite{Garcia2018How}.

\section{Multi-Modal Representations}
\label{sec:encoder}

To jointly represent aesthetics and semantics in art, we propose to project robust visual and language representations enhanced with artistic attributes into a multi-modal artistic space, as depicted in Fig. \ref{fig:model}. In total, we combine four different representations, which are described below.

\paragraph{Language representation} The language representation captures the insights of the high-level semantics of artworks by encoding both titles and artistic comments. Titles are encoded as a term frequency - inverse document frequency (tf-idf) vector, $\vec{v}_{\text{tit}} \in \mathbb{R}^{N_t}$, with $N_t = 9,092$ being the size of the title vocabulary built with the alphabetic words in the titles in the training set. Comments are encoded as another tf-idf vector, $\vec{v}_{\text{com}} \in \mathbb{R}^{N_c}$, with $N_c = 9,708$ being the comments vocabulary built with the alphabetic words occurring at least ten times in the training set. The language representation is obtained by $\vec{v}_\text{lang} = \vec{v}_{\text{tit}} \oplus \vec{v}_{\text{com}}$, where $\oplus$ is vector concatenation. 

\paragraph{Language attributes} Attributes capture the essential information of a painting, such as its painter or its date of creation. We encode the type, school, timeframe, or author labels in the dataset as a one-hot vector, $\vec{v}_{\text{att}} \in \mathbb{R}^c$, with $c$ being the number of labels in each attribute. 

\paragraph{Visual representation} The visual representation captures the visual appearance of paintings. Painting images are scaled down to 256 pixels per side, randomly cropped into 224 $\times$ 224 patches and fed into a ResNet50 \cite{he2016deep}, initialised with its standard pre-trained weights. Appearance is then represented by the output of the model as $\vec{v}_\text{vis} \in \mathbb{R}^{1000}$. 

\begin{figure}
\centering
\includegraphics[width = 0.48\textwidth]{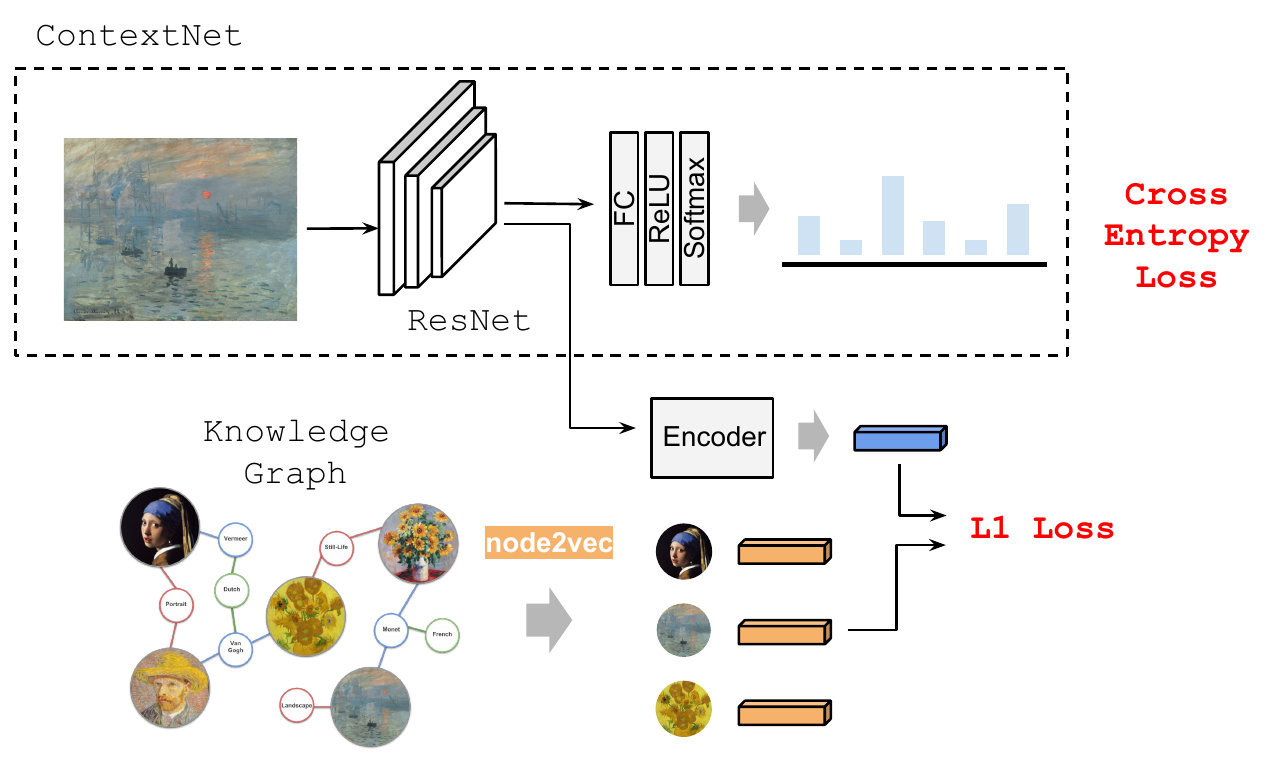}
\caption{ContextNet predicts the painting attributes, such as type, school, timeframe, or author, by fine-tuning a ResNet model based on the information captured by an artistic Knowlegde Graph.} 
\label{fig:kgmodel}
\end{figure}

\paragraph{Image attributes} From the painting image, we use a contextual network (ContextNet) \cite{Garcia2017Context} to predict the artistic attributes. ContextNet is composed by two core modules, as depicted in Fig. \ref{fig:kgmodel}: a ResNet\footnote{Without the last fully connected layer.} \cite{he2016deep}, which obtains the visual information of the image, and a knowledge graph, which captures the contextual relationships of the painting. The visual encoding from the ResNet is further input into an attribute classifier\footnote{A $n$-dimensional fully-connected layer with ReLU and softmax, where $n$ is the number of classes for the predicted attribute.} for predicting the artistic attributes, and into an encoder module\footnote{A 128-dimensional fully-connected layer.} for projecting the visual encoding into the knowledge graph space. The knowledge graph is built by connecting the training paintings in SemArt with their attributes, and its nodes are encoded into a 128-dimensional graph representations using node2vec \cite{grover2016node2vec}.

At training time, we compute the cross-entropy loss function, $\ell_c$, between the predicted attribute and the real attribute of the painting, and the smooth L1 loss function, $\ell_e$, between the encoder output and the graph embedding from the knowledge graph. The ContextNet weights are learnt by jointly optimising both losses as:

\begin{equation}
    \mathcal{L} = \lambda_c \sum_{j=1}^{N} \ell_{c,j} + \lambda_e \sum_{j=1}^{N} \ell_{e,j}
    \label{eq:totalloss}
\end{equation} %
where $\lambda_c$ and $\lambda_e$ are parameters that weight the contribution of the classification and the encoder modules, respectively, and $N$ is the number of training samples. To predict the painting attribute, the graph computation part is removed, and the attribute is predicted as the maximum value of the output of the ContextNet classifier, represented as $\vec{v}_\text{ctx} \in \mathbb{R}^{c}$, with $c$ being the number of labels in the attribute. 

\begin{table*}
\vspace{-3pt}
\centering
\setlength{\tabcolsep}{5pt}
\renewcommand{\arraystretch}{1.1}
\begin{tabular}{ l c c c c c c c c c c}
\hline 
 & \multicolumn{4}{c}{\textbf{Text $\rightarrow$ Image}} & & \multicolumn{4}{c}{\textbf{Image $\rightarrow$ Text}} \\ \cline{2-5} \cline{7-10}

\textbf{Encoding} & \textbf{R@1} & \textbf{R@5} & \textbf{R@10} & \textbf{MR} & & \textbf{R@1} & \textbf{R@5} & \textbf{R@10} & \textbf{MR} \\ \hline 

Vis\&Lang & 0.164 & 0.384 & 0.505 & 10 & & 0.162 & 0.366 & 0.479 & 12 \\

Att \scriptsize{Type} & 0.178 & 0.383 &	0.525 &	9 & & 0.165 &	0.364 &	0.491 &	11 \\
Att \scriptsize{School} & 0.192 & 0.386 & 0.507 & 10 & & 0.163 & 0.364 & 0.484 & 12 \\ 
Att \scriptsize{Tf} & 0.127 & 0.322 & 0.432 & 18 & &	0.130 & 0.336 &	0.444 &	16 \\
Att \scriptsize{Author} & 0.236	& 0.451 & 0.572 & 7 & &	0.204 & 0.440 & 0.535 &	8 \\

Att\&ContextNet \scriptsize{Type} & 0.152 & 0.367 & 0.506 & 10 & & 0.147 & 0.367 & 0.507 & 10\\
Att\&ContextNet \scriptsize{School} & 0.162 & 0.371 & 0.483 & 12 & & 0.156	& 0.355 & 0.483 & 11\\
Att\&ContextNet \scriptsize{Tf} & 0.175 & 0.399 & 0.506 & 10 & & 0.148 & 0.360 & 0.472 & 12 \\
Att\&ContextNet \scriptsize{Author} & \textbf{0.247} & \textbf{0.477} & \textbf{0.581} &	\textbf{6} & & \textbf{0.212} &	\textbf{0.446} &	\textbf{0.563} & \textbf{7}\\
\hline
\end{tabular}
\vspace{3pt}
\caption{Results on the Text2Art Challenge when using vision and language only (Vis\&Lang), when adding attributes (Attributes) and when adding the ContextNet classifier (Att\&ContextNet).}
\label{tab:results}
\end{table*}

\section{Multi-Modal Projections}

To learn the relationship between the visual attributes from the paintings and the semantics from the comments, we project the multi-modal representations from paintings and comments into a multi-modal artistic space. We define the vectors $\vec{p} \in \mathbb{R}^{1000+c}$ and $\vec{q}\in \mathbb{R}^{N_t+N_c+c}$ as the joint representation of visual and image attributes, and language and language attributes, respectively: %
\begin{align*}
\vec{p} = \vec{v}_{\text{vis}} \oplus \vec{v}_{\text{ctx}} \\ 
\vec{q} = \vec{v}_{\text{lang}} \oplus \vec{v}_{\text{att}}
\end{align*}

The two joint representation vectors are projected into a multi-modal artistic space using the non-linear functions $f(\cdot)$ and $g(\cdot)$, respectively, which are implemented with a 128-dimensional fully connected layer followed by a tanh activation function and a $\ell_2$-normalisation layer. The whole model, except for the ContextNet which is previously fine-tuned and frozen, is trained end-to-end using both matching and non-matching pairs of samples from the training set. The loss is computed as a cosine margin loss function: %
\begin{small}
\begin{equation*}
\scriptstyle
\begin{split}\text{Loss}(\vec{p}_i,\vec{q}_j) =
\begin{cases}
1 - \text{sim}(f(\vec{p}_i), g(\vec{q}_j)), & \text{if } i = j \\
\max(0, \text{sim}(f(\vec{p}_i), f_q(\vec{q}_j)) - \Delta), & \text{if } k \neq j
\end{cases}\end{split}
\end{equation*}
\end{small} %
where the sub-indeces $i$ and $j$ are the representations for the $i$-th and $j$-th training sample, $\text{sim}(\cdot, \cdot)$ is the cosine similarity between two vectors, and $\Delta = 0.1$ is the margin. We use Adam optimiser with learning rate 0.0001.
\begin{table}
\vspace{-3pt}
\renewcommand{\arraystretch}{1.1}
\setlength{\tabcolsep}{5pt}
\centering
\begin{tabular}{ l c c}
\hline
\textbf{Model} & \textbf{Easy} & \textbf{Difficult} \\ 
\hline
Vis\&Lang & 0.750 & 0.620\\ 
Att\&Context & 0.830 & 0.680 \\
Human & 0.889 & 0.714\\
\hline
\end{tabular}
\vspace{3pt}
\caption{Multi-modal representations against humans.}
\label{tab:human}
\end{table}

\section{Evaluation}
To evaluate the quality of language and vision representations in art, we design the Text2Art challenge based on multi-modal retrieval, in which the aim is to find the most representative painting given an artistic comment, and vice versa, by ranking test samples according to their cosine similarity. In this way, the challenge evaluates whether the models capture enough of the insights and clues provided by the artistic comments to be able to match it to the correct painting. Results are reported with standard retrieval metrics: median rank (MR), and recall rate at K (R@K), with K being 1, 5 and 10. 

Table \ref{tab:results} reports an ablation study when different combinations of the proposed representations are used. Vis\&Lang uses the visual and language representations only. Att uses the vision, language, and language attribute (specified in brackets) as well as the output of a ResNet152 attribute classification network as a simplier image attribute representation. Note that the image attribute representation predicted in this way has not been informed with the graph representation from the knowledge graph. Finally, Att\&ContextNet considers the four multi-modal representations from Section \ref{sec:encoder}, including the context-aware classifier.

The best results are obtained when the four proposed representations are used, with attributes from language and image are given by the author. Att\&ContextNet (Author) improves results by a 37.24\% in average with respect to vision and language only, suggesting the importance of considering context when studying art. When compared against Att, the use of ResNet152 instead of the context-aware classifier performs better with type and school attributes, whereas  Att\&ContextNet is the best in timeframe and author. 

In Table \ref{tab:human}, we evaluate the proposed multi-modal art representations against human performance, where human evaluators were asked to choose between 10 paintings according to an artistic comment, title, author, type, school, and timeframe. We performed two evaluations: in the easy setup, the 10 paintings were chosen randomly, whereas in the difficult setup, the 10 paintings shared the same type (i.e. landascape, portrait, etc.). The multi-modal representations using the ContextNet reached values close to human accuracy, outperforming Vis\&Lang by a 10.67\% in the easy task and a 9.67\% in the difficult task.

\section{Conclusion}
We addressed art understanding by introducing a new dataset of paintings with associated comments and exploring multi-modal representations in art. Results showed that robust vision and language representations were able to capture the semantic content of paintings relatively well. However, performance was considerably improved when contextual information in the form of a knowledge graph was used to inform the model, which suggested the existence of a strong relationship between art and context. As a future work, we would like to pursue effort in the use of knowledge graph to connect vision and language. We could enhance ContextNet with more robust graph embedding techniques, such as StarSpace \cite{wu2018starspace}, as well as enhance the language representation with the knowledge graph attributes. 

\textbf{Acknowledgement}: This work was partly supported by JSPS KAKENHI Grant No.~18H03264.

{\small
\bibliographystyle{ieee}
\bibliography{bibliography}
}

\end{document}